\documentclass[letterpaper, 10 pt, conference]{ieeeconf}  

\IEEEoverridecommandlockouts                              

\usepackage{graphics} 
\usepackage{epsfig} 
\usepackage{mathptmx} 
\usepackage{times} 
\usepackage{amsmath} 
\usepackage{amssymb}  
\usepackage{cite}
\usepackage[dvipsnames]{xcolor}
\usepackage{booktabs}
\usepackage{float}
\usepackage{multirow}
\usepackage{stfloats}
\usepackage{algorithm}
\usepackage{algpseudocode}  
\usepackage{caption}
\usepackage{cuted}
\usepackage{newtxtext, newtxmath}
\usepackage{hyperref}
\usepackage{balance}

\usepackage{array}        
\usepackage{xcolor}       
\usepackage{colortbl}
\definecolor{grey}{rgb}{0.9, 0.9, 0.9}

\usepackage{graphicx}
\usepackage{subcaption} 
\usepackage{caption}    
\usepackage{lipsum}     
\usepackage{array}        
\usepackage{xcolor}       
\usepackage{colortbl}
\definecolor{ourcolor}{HTML}{99e0eb}
\definecolor{ourblue}{HTML}{27a2c3}

\definecolor{tablecolor}{HTML}{ccf2f5} 

\definecolor{tablecolor2}{HTML}{ffcdb4}
\definecolor{citecolor}{HTML}{fe7b5b}
\definecolor{grey}{rgb}{0.9, 0.9, 0.9}
\usepackage{amssymb}

\title{\fontsize{18}{2}\selectfont\bfseries
Factor-Aware Mixture-of-Experts with Pretrained Encoder \\[8pt]
\fontsize{18}{2}\selectfont for Combinatorial Generalization}

\author{Feihong Zhang$^{1}$, Guojian Zhan$^{1}$, Zeyu He$^{1}$, Yinuo Wang$^{1}$, Likun Wang$^{1}$, Tianze Zhu$^{1}$,\\ Yao Lyu$^{1}$, Tao Zhang$^{2}$, Tinghao Yi$^{3}$, Wei You$^{3}$, Shengbo Eben Li†$^{1}$
\thanks{This study is supported by Tsinghua University-EFORT Intelligent Equipment Co., Ltd Joint Research Center for Embodied Intelligence Computing and Perception. This work is also supported by SunRisingAI Lab.}
\thanks{$^{1}$ Tsinghua University, Beijing, China}%
\thanks{$^{2}$ SunRisingAI Ltd., Beijing, China}%
\thanks{$^{3}$ EFORT Intelligent Robot Co., Ltd., Anhui, China}%
\thanks{† Corresponding author. E-mail: \texttt{lishbo@tsinghua.edu.cn}}%
}

\begin{document}
\maketitle
\thispagestyle{empty}
\pagestyle{empty}


\begin{abstract}

The integration of pretrained encoders with diffusion policies has become a dominant paradigm for visual robotic manipulation. However, it still struggles to generalize across complex environments with varying factors such as lighting and surface textures.
To address this, we propose FAME, a framework that integrates a factor-aware mixture-of-experts (MoE) with a pretrained encoder to enhance generalization to environmental variations. FAME follows a three-stage training process: (1) policy warmup, where a diffusion policy is trained on standard-environment data with a frozen encoder; (2) factor-specific adapter training, where lightweight adapters inserted between the frozen encoder and the temporarily frozen policy are trained on customized datasets, each targeting a distinct environmental variation; and (3) joint fine-tuning, where a central router and the warmed policy are trained on mixed data to handle multiple factors jointly. FAME is ``factor-aware'' because the central router softly weights frozen factor-specific adapters as a dense MoE, enabling combinatorial generalization across multiple factors.
Evaluations on the Meta-World benchmark show that FAME outperforms diffusion policy baselines by 34\%. We further validate FAME in a real-world pick-and-place task using a compact model trained on newly collected data, where FAME achieves a 35\% improvement in generalization under real-world variations.

\end{abstract}

\section{INTRODUCTION}
\label{sec:introduction}

The adoption of Diffusion Policies (DP) ~\cite{chi2023diffusion_policy, shao2026x} has become a well-established consensus in visual robotic manipulation, owing to their powerful fitting capabilities for complex, high-dimensional tasks. This has led to the prevailing approach of integrating DP with various pre-trained visual encoders, which provides rich, transferable feature representations. Representative encoders include DINOv2~\cite{oquab2023dinov2}, CLIP~\cite{radford2021learning} and R3M~\cite{nair2022r3m}. 

\begin{figure}[htbp]
\centering
\includegraphics[
width=0.5\textwidth,
trim=8.0cm 1.0cm 2.2cm 1.0cm,  
clip
]{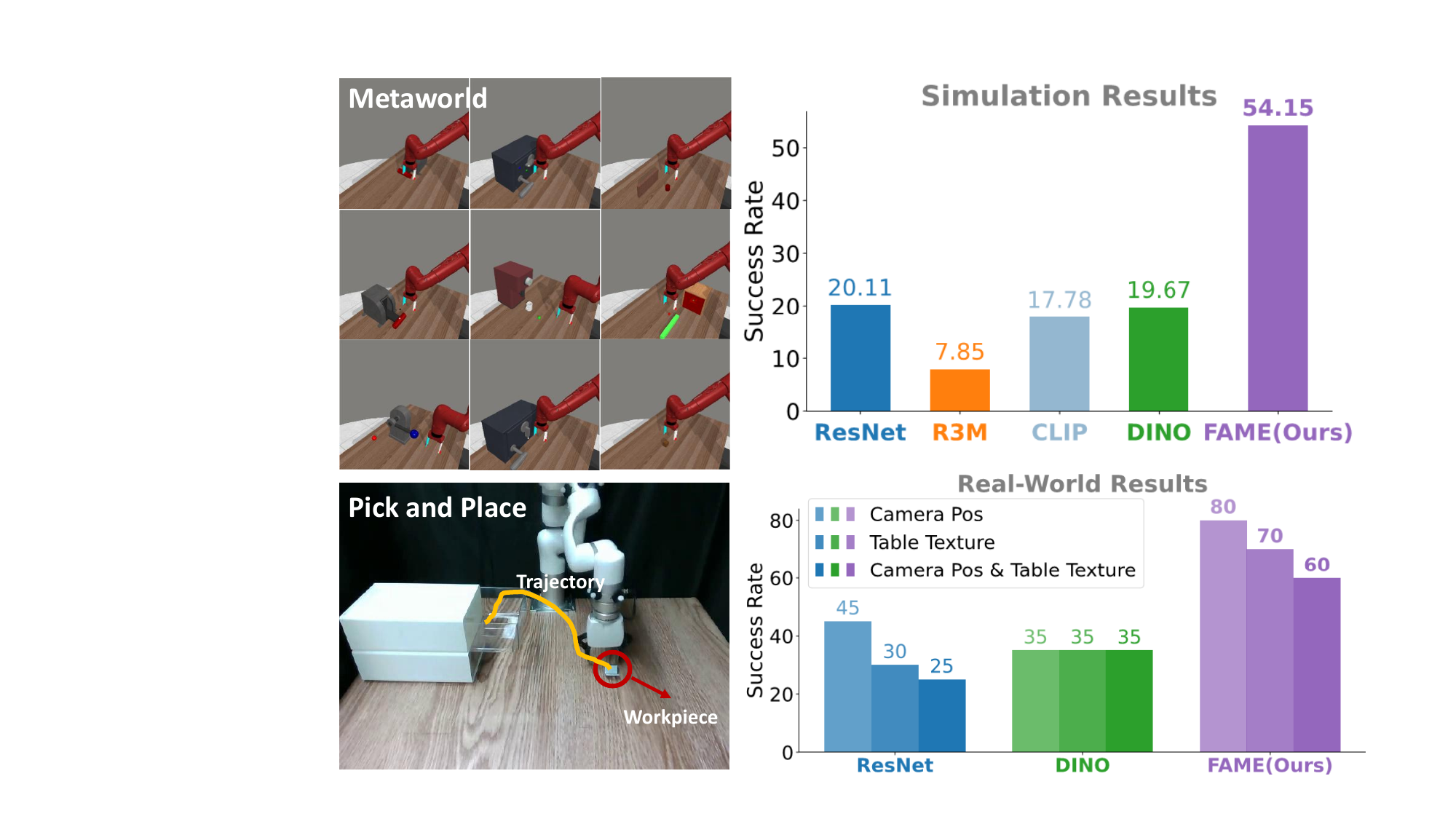}
\caption{FAME is a novel visual motor learning framework for combinatorial generalization. It achieves best performance surpassing baseline methods in both MetaWorld simulation and real-world experiments.}
\label{fig:introduction}
\end{figure}




Despite these advancements, current methods still struggle to generalize across complex environments with varying factors such as lighting, surface textures, or camera position. If mastering each factor requires additional data of size $N$, then simultaneously handling $K$ independent factors could imply a considerable data complexity of $N^K$, which becomes prohibitively expensive in real world. 

Fortunately, many factors in the physical world vary independently. This observation motivates a divide-and-conquer strategy: by disentangling and separately learning each factor, we can potentially reduce the data requirement from exponential to approximately linear, i.e., $N \times K$. Real-world environmental changes can be decomposed into discrete and independent factors. Explicitly modeling these variations enables more systematic and combinatorial adaptation to a majority of common conditions.

Furthermore, directly fine-tuning large pre-trained encoders remains challenging: it is computationally expensive, prone to overfitting, and often results in catastrophic forgetting of pre-trained knowledge. To overcome these limitations, we propose a structured approach that factorizes environmental variations, enabling efficient and scalable combinatorial generalization in complex visual manipulation tasks.

In this paper, we introduce \textbf{FAME} (\underline{F}actor-\underline{A}ware \underline{M}ixture-of-Experts with Pretrained \underline{E}ncoder), a novel framework that enhances the generalization capability of diffusion policies through factor-aware adaptation. FAME incorporates a dense Mixture-of-Experts (MoE) architecture that softly combines lightweight, factor-specific adapters, each dedicated to a specific environmental variation. The training process consists of three stages: (1) \textbf{Policy warm-up}: A diffusion policy is first trained using a frozen pretrained encoder on data from a standard environment. (2) \textbf{Factor-specific adapter training}: Lightweight adapters are inserted between the encoder and the policy network and trained separately on specialized datasets, each targeting a distinct environmental factor. (3) \textbf{Joint fine-tuning}: A central router is trained along with the policy on a mixed dataset to combine adapters dynamically and achieve combinatorial generalization.


Our contributions are summarized as follows:
\begin{itemize}
\item We propose FAME, a factor-aware framework that integrates a Mixture-of-Experts(MoE) architecture with a frozen pretrained encoder to handle compound environmental variations in visual robotic manipulation.
\item We design a three-stage training procedure that includes policy warm-up, factor-specific adapter training, and joint fine-tuning with a router, enabling efficient and scalable adaptation.
\item Extensive experiments on both simulated benchmarks and real world validate FAME's effectiveness, achieving a 34\% performance gain in simulation and a 35\% improvement in real-world generalization over baselines.
\end{itemize}

\section{RELATED WORKS}
\label{sec:related_work}

\vspace{0.5em}

\subsection{Diffusion policy and robotic manipulation.}

Diffusion models, which progressively transform random noise into structured data samples, have demonstrated remarkable success in high-fidelity image generation, as exemplified by {DDPM} \cite{ho2020ddpm, song2020score}. Owing to their strong representational power, such models are increasingly being adopted in robotics. For instance, they have been applied in reinforcement learning~\cite{wang2024diffusion, li2025hybrid, sheng2026mp1}, and in imitation learning~\cite{chi2023diffusion_policy, huang2025memory, tie2025seed}. In this work, we focus on leveraging diffusion models for robotic manipulation under complex generalization scenarios. We investigate how diffusion-based policies, formulated as conditional diffusion models, can be improved through architectural modifications to enhance the generalization capability of robotic policy learning.
    
\subsection{Pre-trained visual encoders.}

In the realm of computer vision, several prominent pre-trained visual encoders have emerged as powerful feature extractors, including {Vision Transformer (ViT)}~\cite{dosovitskiy2021an}, {DINOv2}~\cite{oquab2023dinov2}, and {CLIP}~\cite{radford2021learning}. Among these, DINOv2—a robust visual encoder based on self-supervised learning—has been extensively applied in embodied motion vision due to its strong representation capabilities. These general-purpose encoders have subsequently inspired and facilitated the development of specialized visual encoders within the field of robotic policy learning. Notable contributions include {MVP}~\cite{xiao2022masked}, {R3M}~\cite{nair2022r3m}, {VIP}~\cite{ma2022vip}, and {VC-1}~\cite{majumdar2024we}, which leverage large-scale pre-training to provide effective visual representations that serve as valuable prior knowledge for training robot policies. In this paper, we employ the pre-trained visual representations from DINOv2~\cite{oquab2023dinov2} and our framework is compatible to any other encoders.

\subsection{Parameter-efficient fine-tuning.}

Despite the strong representational capabilities of pre-trained visual encoders, their limited adaptability to environmental variations poses a significant challenge for robotic manipulation. To address this issue, we draw inspiration from {Parameter-Efficient Fine-Tuning (PEFT)} methods developed in natural language processing. Instead of full fine-tuning that updates all parameters, these approaches introduce small trainable modules into frozen pre-trained backbones, preserving the original representations while enabling task-specific adaptation. Seminal work in this area includes {Adapter} modules~\cite{houlsby2019parameter} and {Low-Rank Adaptation (LoRA)}~\cite{hu2021lora}, alongside other techniques like {Prompt Tuning}~\cite{lester2021power} and {Prefix Tuning}~\cite{li2021prefix}. These methods have demonstrated remarkable success in adapting large language models with minimal computational overhead. Our work extends this parameter-efficient paradigm to visual representation learning for robotic manipulation, developing factor-specific adapters that maintain the benefits of large-scale pre-trained visual encoders while enabling efficient adaptation to diverse environmental conditions.

\vspace{0.3em}

\subsection{Mixture-of-Experts (MoE) frameworks.}

The MoE architecture provides an effective mechanism for dynamically integrating multiple specialized modules. Originally introduced by \cite{shazeer2017outrageously}, MoE enables scalable neural networks by selectively routing inputs to specialized "expert" sub-networks. This approach has demonstrated remarkable success in large language models, including the {Switch Transformer} \cite{fedus2021switch} and {Mixtral 8x7B} \cite{jiang2024mixtral}. Beyond natural language processing, MoE has been effectively applied in {autonomous driving} \cite{sun2025generalizing, yang2026drivemoe}, as well as in {robotics} \cite{song2024germ, cheng2025moedpmoeenhanceddiffusionpolicy}. Our work combines parameter-efficient adaptation and mixture-of-experts by developing a FAME framework that softly integrates factor-specific adapters. Unlike sparse-routing MoE, FAME uses dense weighted aggregation to combine specialized adapters under compound environmental contexts, addressing combinatorial generalization in robotic manipulation scenarios.

\section{METHOD}
\label{sec:method}

\vspace{0.5em}

In this section, we introduce the core methodology of the FAME framework. This framework addresses the challenge of diverse environmental variations in robotic manipulation by combining a three-phase training approach with a dynamic MoE mechanism.

\subsection{Overview of FAME framework}
\label{subsec:factor_aware_adapter_training}

\begin{figure*}[!t]
    \centering
    \includegraphics[width=\linewidth, trim=5mm 3mm 5mm 3mm, clip]{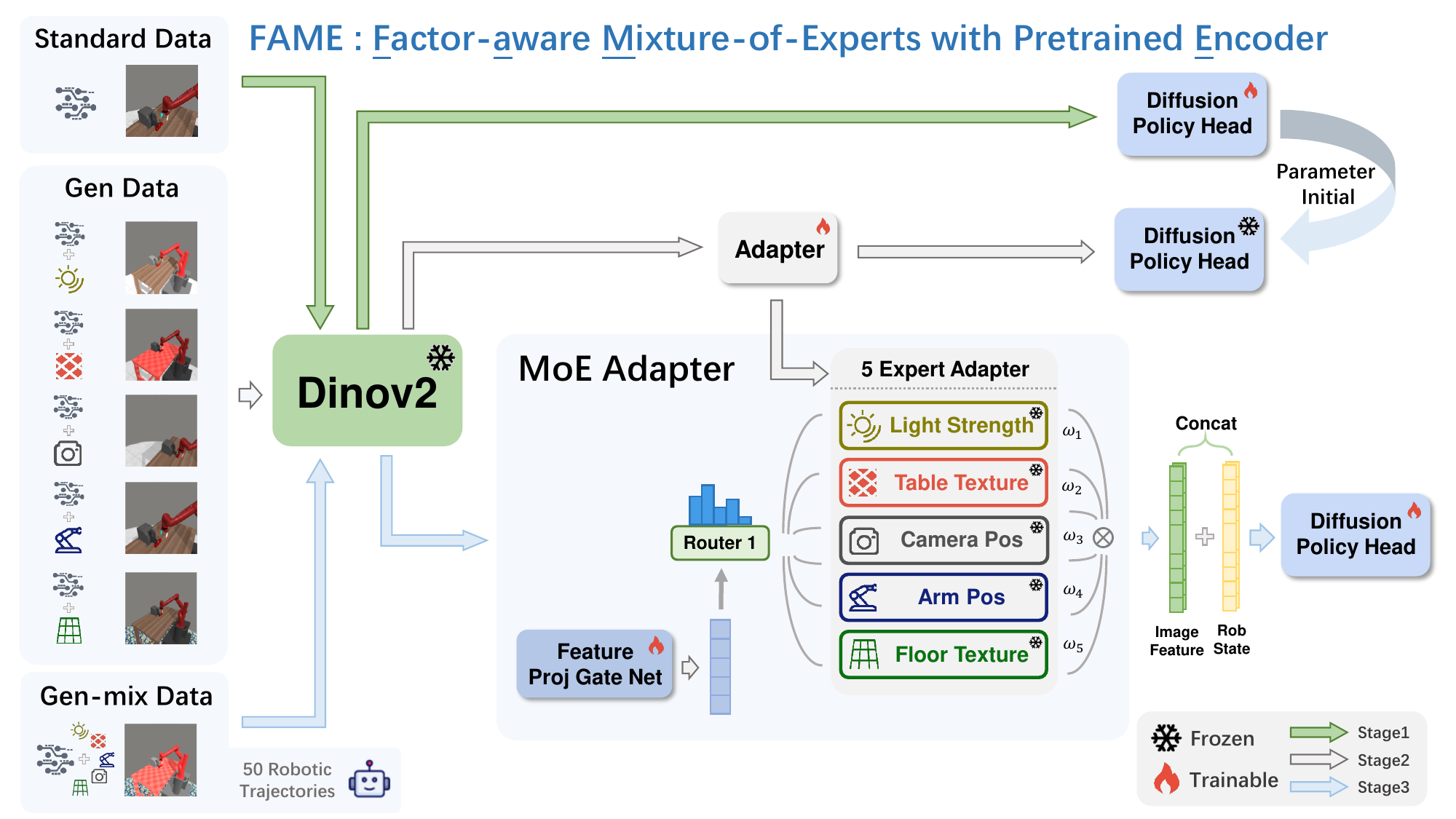}
    \caption{FAME framework: (1) Policy warm-up: The standard DP framework serves as the baseline policy training; (2) Factor-specific adapter training: Multiple adapters are trained on a frozen DINOv2 backbone to handle individual variations (e.g., lighting, texture); (3) Joint fine-tuning: A gating network dynamically combines adapter outputs via Mixture-of-Experts.}
    \label{fig:FAME_Framework}
\end{figure*}

The framework of our FAME is illustrated in Figure~\ref{fig:FAME_Framework}, where the training process is summarized using color-coded arrows: green arrows denote policy warm-up (Phase 1 in Section~\ref{subsec:phase1}), in which a diffusion policy is trained using a frozen pretrained encoder on data from a standard environment; gray arrows represent factor-specific adapter training (Phase 2 in Section~\ref{subsec:phase2}), where lightweight adapters are inserted and trained separately on specialized datasets, each targeting a distinct environmental factor; and blue arrows correspond to joint fine-tuning (Phase 3 in Section~\ref{subsec:phase3}), during which a central router is trained along with the policy on a mixed dataset to combine adapters dynamically.

Before detailing the model architecture and training procedures in the following subsections, we first introduce the three types of datasets used across different stages of the training framework:

(1) \textbf{Standard Dataset ($\mathcal{D}_0$)}: Data collected in the standard manipulation task environment.

(2) \textbf{Gen Dataset ($\mathcal{D}_k$)}: Data collected under environments where only the $k$-th factor (e.g., light strength) is varied relative to the standard setup, for each $k \in {1,\dots,K}$.

(3) \textbf{Mix Gen Dataset ($\mathcal{D}_{\text{multi}}$)}: Data collected under environments where any subset of $i$ factors vary simultaneously, with $i \in \{2,3,4,K\}$.

\subsection{Phase 1: Policy warm-up}
\label{subsec:phase1}
The first phase aims to learn a base policy that performs well under standard environmental conditions.

We adopt the two-stage end-to-end diffusion policy (DP) architecture as the backbone of our framework. The first stage employs a visual backbone based on a frozen pre-trained DINOv2 \cite{oquab2023dinov2} model to leverage its powerful representation capabilities. The second stage consists of a diffusion policy head which is trained from scratch. Training uses standard task data $\mathcal{D}_{0}$ without environmental variations. Given input observation $\mathbf{o}_t^{0}$ (where the top-right label "$0$" represents the dataset to which $\mathbf{o}_t$ belongs), the visual backbone extracts features $\mathbf{f}_v = \mathcal{H}^{\scalebox{0.6}{\text{frozen}}}{\scalebox{0.5}{\text{DINOv2}}}(\mathbf{o}_t^{0})$, and the DP head generates actions $\mathbf{a}_t = \mathcal{H}{\scalebox{0.5}{\text{DP}}}(\mathbf{f}_v)$. The training objective is to minimize the loss function:
\begin{equation}
\min_{\theta_{\mathcal{H}_{\scalebox{0.5}{\text{DP}}}}} \mathcal{L}_{\scalebox{0.5}{\text{DP}}}(\mathcal{D}_{0}; \theta_{\mathcal{H}_{\scalebox{0.5}{\text{DP}}}}, \mathcal{H}^{\scalebox{0.6}{\text{frozen}}}_{\scalebox{0.5}{\text{DINOv2}}}),
\end{equation}
where $\theta_{\mathcal{H}{\scalebox{0.5}{\text{DP}}}}$ denotes the parameters of the diffusion policy. This phase establishes a strong baseline policy that performs well under standard environmental conditions.

\subsection{Phase 2: Factor-specific adapter training}
\label{subsec:phase2}

In the second phase, we train specialized adapter networks for each environmental factor while keeping both the visual backbone and the DP head frozen.

For each environmental factor $k \in \{1,\dots,K\}$, we introduce a trainable adapter network $\mathcal{A}_k$ between the frozen DINOv2 and the frozen DP head obtained from Phase 1. The visual features $\mathbf{f}_v^{'}$ are first extracted by the frozen DINOv2 model as $\mathbf{f}_v^{'} = \mathcal{H}^{\scalebox{0.6}{\text{frozen}}}_{\scalebox{0.5}{\text{DINOv2}}}(\mathbf{o}_t^{k})$, where $\mathbf{o}_t^{k}$ denotes the input observations from dataset $\mathcal{D}_k$. The adapter network $\mathcal{A}_k$ then transforms these features into adapted visual features $\mathbf{f}_v^{k} = \mathcal{A}_{k}(\mathbf{f}_v^{'})$, which are passed through the frozen DP head to obtain the output $\mathbf{a}_t = \mathcal{H}_{\scalebox{0.5}{\text{DP}}}^{\scalebox{0.6}{\text{frozen}}}(\mathbf{f}_v^{k})$. The training objective for the adapter $\mathcal{A}_k$ is to minimize the loss function $\mathcal{L}_{\scalebox{0.5}{\text{DP}}}$ with respect to the adapter's parameters $\theta_{\mathcal{A}_k}$, while keeping the DINOv2 and DP head models frozen:
\begin{equation}
\min_{\theta_{\mathcal{A}_k}} \mathcal{L}_{\scalebox{0.5}{\text{DP}}}(\mathcal{D}_k; \theta_{\mathcal{A}_k}, \mathcal{H}^{\scalebox{0.6}{\text{frozen}}}_{\scalebox{0.5}{\text{DINOv2}}}, \mathcal{H}^{\scalebox{0.6}{\text{frozen}}}_{\scalebox{0.5}{\text{DP}}}), k \in \{1,\dots,K\}.
\end{equation}
This formulation ensures that each adapter $\mathcal{A}_k$ learns to adapt the visual features specifically for the variations present in dataset $\mathcal{D}_k$, effectively specializing in handling a particular environmental factor while maintaining the base policy's core functionality.

\subsection{Phase 3: Joint fine-tuning}
\label{subsec:phase3}

The final phase integrates the specialized adapter networks through a dense MoE architecture, enabling dynamic combination of expert representations based on input conditions. The gating mechanism learns soft weights for all adapters rather than sparse expert selection. The MoE layer comprises two components:


1. \textbf{Gating network} $\mathcal{G}$: This network learns to compute adapter weights $\mathbf{w} = [w_1,\dots,w_k,\dots ,w_K]$ from the visual features $\mathbf{f}_v^{'}$. The gating network essentially acts as a router, determining the contribution of each expert based on the input characteristics.

2. \textbf{Adapter bank}: This includes pre-trained factor-specific adapter networks $\mathcal{A}_k$ for $k \in \{1,\dots,K\}$ from Phase 2 in Section~\ref{subsec:phase2}, which remain frozen during the MoE training process. These feature-level adapters serve as specialized experts, each handling a specific environmental variation.

The final visual representation is obtained by combining the outputs of the adapter networks via a weighted summation:
\begin{equation}
\mathbf{f}^{\scalebox{0.5}{\text{MoE}}}_v = \sum_{k=1}^{K} \underbrace{\text{Softmax}\left( \mathcal{G}(\mathbf{f}_v^{'}) \right)}_{w_k} \cdot \underbrace{\mathcal{A}_k(\mathbf{f}_v^{'})}_{\mathbf{f}_{v}^{k}}
\end{equation}
This combined visual representation $\mathbf{f}_v^{\scalebox{0.5}{\text{MoE}}}$ is then passed through the DP head to produce the final output: $\mathbf{a}_t = \mathcal{H}_{\scalebox{0.5}{\text{DP}}}(\mathbf{f}_v^{\scalebox{0.5}{\text{MoE}}})$.

\vspace{0.5em}

\textbf{Training procedure.} During training, we utilize multi-factor variation data $\mathcal{D}_{\text{multi}}$ to optimize only the gating Network $\mathcal{G}$ and a new DP head, while keeping the visual backbone and all adapter networks frozen. This training strategy allows the gating network to learn effective combination strategies while preventing catastrophic forgetting of the specialized adapter capabilities. The specific training objective is
\begin{equation}
\min_{\theta_{\mathcal{G}}, \theta_{\mathcal{H}_{\scalebox{0.5}{\text{DP}}}}} \mathcal{L}_{\scalebox{0.5}{\text{DP}}}\left( \mathcal{D}_{\text{multi}}; \theta_{\mathcal{G}}, \theta_{\mathcal{H}_{\scalebox{0.5}{\text{DP}}}}, \mathcal{H}^{\scalebox{0.6}{\text{frozen}}}_{\scalebox{0.5}{\text{DINOv2}}}, \mathcal{A}_k^{\scalebox{0.6}{\text{frozen}}} \right) , \quad k \in \{1,\dots,K\}
\end{equation}
Our framework enables the agent to dynamically adapt to complex environmental conditions by intelligently combining the specialized knowledge of multiple experts, resulting in robust performance across diverse scenarios.

\vspace{0.3em}

\section{EXPERIMENT}
\label{sec:result}

\vspace{0.5em}

\subsection{Simulation Experiment}

\subsubsection{Meta-World benchmark} Meta-World benchmark \cite{yu2020meta} is a widely recognized platform for robotic manipulation that provides a diverse set of tasks simulating real-world scenarios. We choose a representative subset of \textbf{9 tasks} from this benchmark to conduct experiments.


\subsubsection{Environment customization} Meta-World provides only the standard environment interface without variations. To enable our research on generalization, we  develop \texttt{MetaWorldEnvFactor}, a lightweight wrapper class that can be directly nested on top of the original \texttt{MetaWorldEnv}. We implement 5 independent factor variations (object size, color, lighting, friction, and camera pose) and can arbitrarily compose them to customize environments with diverse factor combinations.

\subsubsection{Training dataset}
We use Meta-World’s built-in policies to construct dataset.  By iterating the inference-execution loop until success, high-quality expert trajectories (image-state-action sequences) are collected as the Standard Dataset ($\mathcal{D}_0$). With the help of \texttt{MetaWorldEnvFactor}, we can further construct the Gen Dataset ($\mathcal{D}_k$) and Mix Gen Dataset ($\mathcal{D}_\text{multi}$). Each dataset contains 50 successful demonstrations.



\subsubsection{Evaluation setting} To thoroughly evaluate the policy robustness, we build 5 test environments for each task. Using Hand-Pull as a representative example, the 5 test environments exhibit progressively increasing complexity, ranging from single-factor variations to the most challenging scenario with all five factors simultaneously involved. Evaluation is performed every 200 epochs, resulting in 10 evaluations over the 2000-epoch training run. In each evaluation round, the model is assessed in all 5 test environments ($i=1,2,3,4,5$), yielding 5 individual results. The average of these 5 results is then taken as the evaluation outcome for that particular round.


\subsubsection{Baselines} We consider several well-known method in visual robotic manipulation, including DP with ResNet~\cite{he2016deep}, DP with DINOv2~\cite{oquab2023dinov2} (a ViT-based encoder that learns high-quality visual representations via self-supervised pre-training on large-scale unlabeled image data), DP with CLIP~\cite{radford2021learning} (a vision-language model trained on a massive web-scale dataset of image-text pairs), and DP with R3M~\cite{nair2022r3m} (self-supervised pre-training on large-scale human video data).
Our FAME-DP also employs DP as the downstream controller, while the major difference is that we design a factor-aware MoE to collaborate with DINO encoder for better combinatorial generalization capability.

\subsubsection{Main results}
\begin{figure*}[t]
    \centering
    \includegraphics[width=1.0\linewidth, trim=0mm 0mm 0mm 0mm, clip]{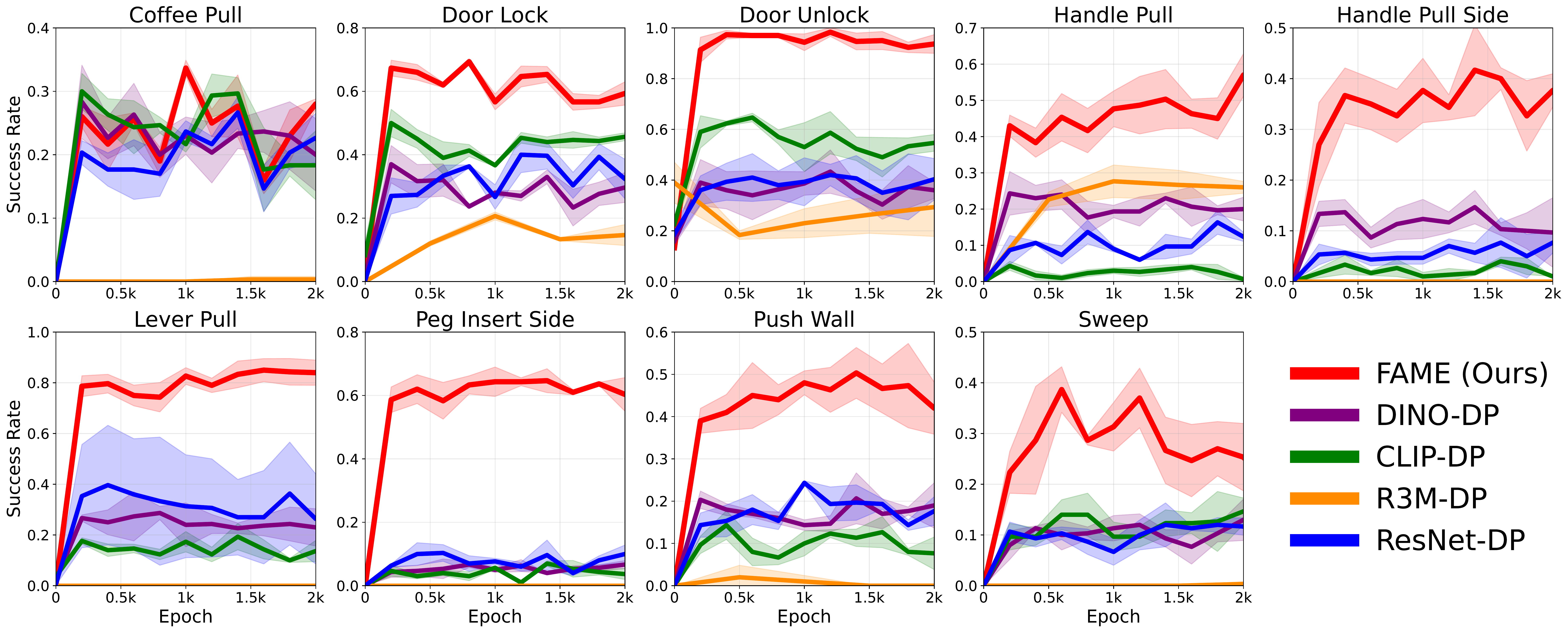}
    \caption{\textbf{Training curves on benchmarks.} The solid lines correspond to the mean and shaded regions correspond to one standard deviation over three runs. Each evaluation result is averaged across five environments with $i=1$, $i=2$, $i=3$, $i=4$, and $i=5$ varying factors.}
    \label{fig:moe_results_comparison}
\end{figure*}

For each task, we run 3 random seeds and each evaluation result is the average outcome across 5 test environments with 1 to 5 varying factors ($i=1,2,3,4,5$). The numerical results of all algorithms are summarized in Table~\ref{table:metaworld results} and the curves are shown in Figure~\ref{fig:moe_results_comparison}. Our approach consistently achieves the highest performance, with an average success rate of $\mathbf{54.15\%}$ across all environmental settings, surpassing the second-best method by 34\% over 9 tasks. Notably, on challenging tasks such as \textbf{Door Lock}, \textbf{Handle Pull Side}, and \textbf{Peg Insert Side}, our method outperforms all baselines by a large margin—achieving $59.33\%$, $37.67\%$, and $60.33\%$ respectively. Furthermore, FAME excels in tasks like \textbf{Door Unlock} and \textbf{Lever Pull}, reaching success rates of $93.67\%$ and $84.00\%$, significantly higher than other methods. These results affirm the strong generalization capability of FAME when faced with diverse and unseen environmental variations.

\begin{table*}[h]
\centering
\caption{\textbf{Average final success rate.} We report the mean $\pm$ one standard deviation over three random seeds of the evaluation results obtained at the 2000\textsuperscript{th} epoch.}
\label{table:metaworld results}

\small
\begin{tabular}{lcccccc}
\toprule
\textbf{Task} & \textbf{ResNet-DP} & \textbf{R3M-DP} & \textbf{CLIP-DP} & \textbf{DINO-DP} & \textbf{FAME (Ours)} \\
\midrule
Coffee Pull & $22.67 \pm 3.86$ & $0.33 \pm 0.47$ & $18.33 \pm 5.44$ & $20.00 \pm 5.72$ & $\mathbf{28.00 \pm 0.82}$ \\
Door Lock & $32.33 \pm 6.24$ & $14.67 \pm 3.30$ & $45.67 \pm 1.25$ & $29.67 \pm 4.71$ & $\mathbf{59.33 \pm 3.68}$ \\
Push Wall & $17.67 \pm 3.40$ & $0.00 \pm 0.00$ & $7.67 \pm 3.86$ & $19.00 \pm 5.35$ & $\mathbf{42.00 \pm 6.16}$ \\
Sweep & $11.67 \pm 1.70$ & $0.33 \pm 0.47$ & $14.67 \pm 2.62$ & $13.00 \pm 4.08$ & $\mathbf{25.33 \pm 6.65}$ \\
Lever Pull & $26.33 \pm 17.75$ & $0.00 \pm 0.00$ & $13.67 \pm 4.19$ & $23.00 \pm 7.35$ & $\mathbf{84.00 \pm 4.97}$ \\
Door Unlock & $40.33 \pm 8.18$ & $29.33 \pm 11.56$ & $54.67 \pm 3.30$ & $36.00 \pm 3.27$ & $\mathbf{93.67 \pm 3.68}$ \\
Handle Pull & $12.33 \pm 1.25$ & $26.00 \pm 1.63$ & $0.67 \pm 0.47$ & $20.00 \pm 3.27$ & $\mathbf{57.00 \pm 5.89}$ \\
Handle Pull Side & $7.67 \pm 2.05$ & $0.00 \pm 0.00$ & $1.00 \pm 0.00$ & $9.67 \pm 6.85$ & $\mathbf{37.67 \pm 3.30}$ \\
Peg Insert Side & $10.00 \pm 2.83$ & $0.00 \pm 0.00$ & $3.67 \pm 1.70$ & $6.67 \pm 1.25$ & $\mathbf{60.33 \pm 5.31}$ \\
\midrule
\rowcolor{gray!20}
Average & $20.11$ & $7.85$ & $17.78$ & $19.67$ & $\mathbf{54.15}$ \\
\bottomrule
\end{tabular}
\end{table*}

\subsection{Simulation Ablation Study}

To investigate the core properties of our FAME framework, we conducted a detailed ablation study on the \textbf{Handle-Pull} task. 

\subsubsection{Scaling property with data increasing} 
\begin{figure}[htbp]
    \centering
    \includegraphics[
        width=0.45\textwidth,
        trim=0cm 0cm 0cm 0cm, 
        clip 
    ]{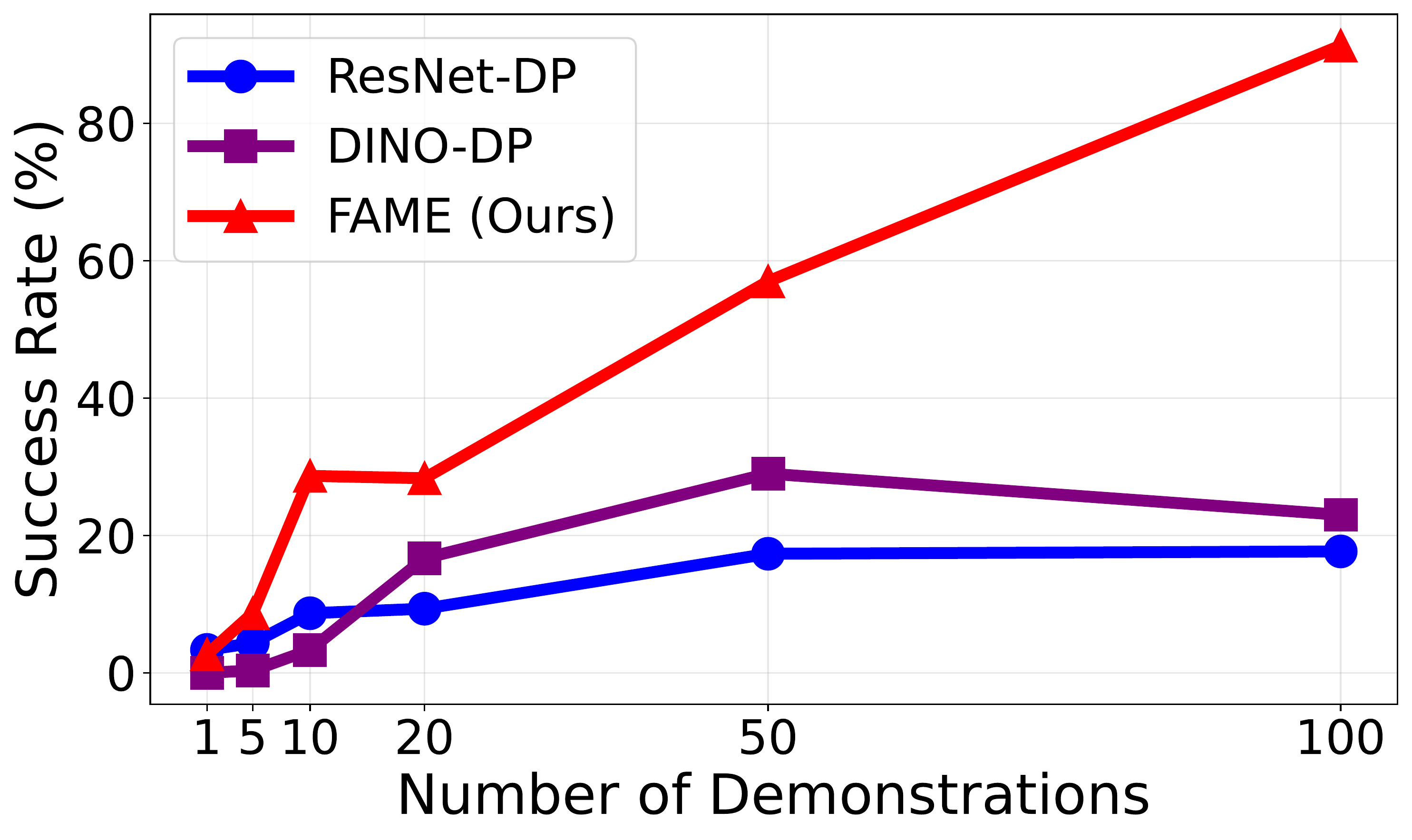}
    \caption{\textbf{Scaling performance with increasing demonstration data.} Evaluation of FAME and baselines trained on the Mix Gen Dataset ($\mathcal{D}_{\text{multi}}$) with varying numbers of demonstrations.}
    \label{fig:handle_pull_success_rate}
\end{figure}

We evaluated the scaling effects of our FAME framework by training on varying dataset sizes (1, 5, 10, 20, 50, and 100 demonstrations), using the same Mix Gen Dataset ($\mathcal{D}_{\text{multi}}$, $i=5$) as in the main experiments. As shown in Figure \ref{fig:handle_pull_success_rate}, our algorithm consistently outperformed baselines across all scales. The results reveal a strong scaling behavior, with performance improving significantly as data volume increases. This demonstrates that our framework effectively leverages larger datasets to enhance generalization, a key advantage that highlights the effectiveness of combining a pre-trained encoder with a dynamic MoE structure.

\subsubsection{Performance considering only single factor variation at a time} 
\begin{figure}[htbp]
    \centering
    \includegraphics[
        width=0.50\textwidth, 
        trim=0cm 0cm 0cm 0cm, 
        clip 
    ]{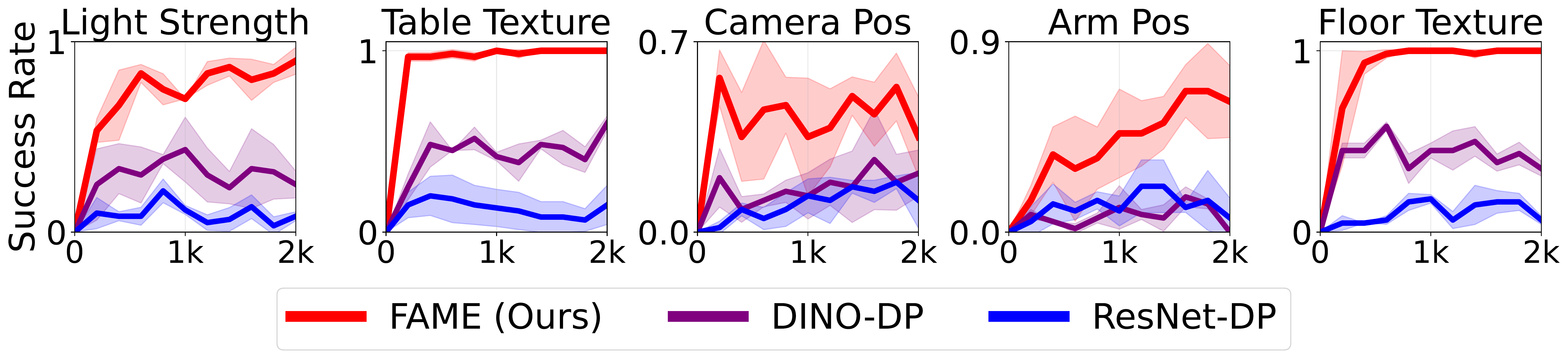}
    \caption{ Evaluation on environments containing only one varying factor at a time (Gen Dataset $\mathcal{D}_k$).}
    \label{fig:ablation_results_comparison}
\end{figure}

While our main experiments showed strong performance on multi-factor variations, we also evaluated our FAME framework's ability to handle single-factor changes. For this, we trained and evaluated the model using only the Gen Dataset ($\mathcal{D}_k$), where {each environment contained a single varying factor}. As shown in Figure \ref{fig:ablation_results_comparison}, our FAME algorithm maintains strong performance across all five individual factors. This demonstrates the framework's robust adaptability, proving it is highly effective at addressing both single-factor and multiple-factors environmental challenges. For consistency and fair comparison, we utilize the smallest available parameter size versions for all pre-trained models mentioned above.

\subsubsection{Zero-shot Cross-Task Generalization of the Gating Network in FAME}

\begin{figure}[htbp]
    \centering
    \subfloat[HP-HP\label{subFig:1}]{\includegraphics[
        width=0.24\textwidth, 
        trim=0.0cm 10.0cm 25.0cm 1.0cm, 
        clip 
    ]{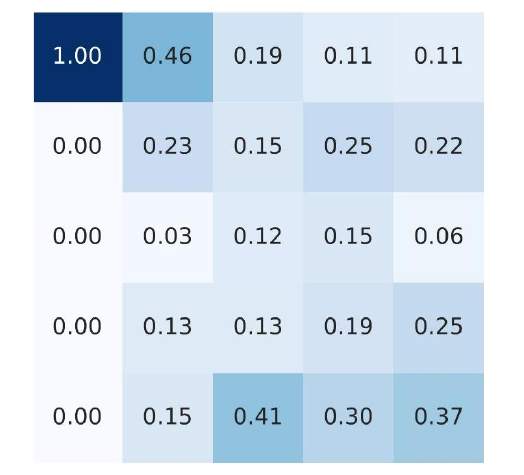}}
    \subfloat[PIS-PIS\label{subFig:2}]{\includegraphics[
        width=0.24\textwidth, 
        trim=0.0cm 10.0cm 25.0cm 1.0cm, 
        clip 
    ]{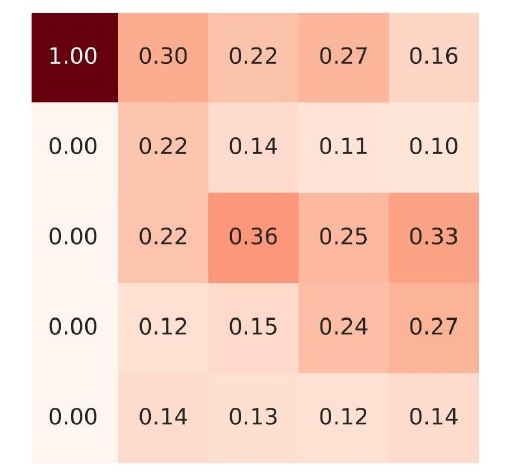}}
    \\
    \subfloat[HP-PIS\label{subFig:3}]{\includegraphics[
        width=0.24\textwidth, 
        trim=0.0cm 10.0cm 25.0cm 1.0cm, 
        clip 
    ]{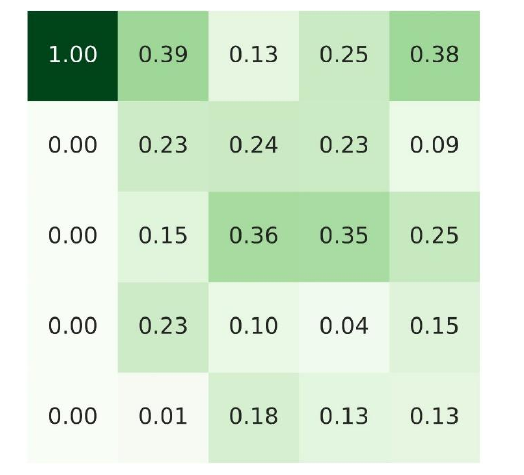}}
    \subfloat[PIS-HP\label{subFig:4}]{\includegraphics[
        width=0.24\textwidth, 
        trim=0.0cm 10.0cm 25.0cm 1.0cm, 
        clip 
    ]{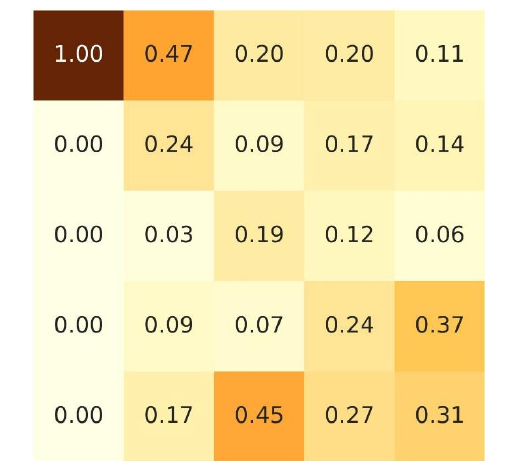}}
    \caption{\textbf{Cross-task generalization of the gating network in FAME.} Heatmaps show average router weights on Handle Pull (HP) and Peg Insert Side (PIS). Rows correspond to five factor-specific experts in the order of Light Strength, Table Texture, Camera Pose, Arm Pose, and Floor Texture; columns correspond to environments with 1 to 5 mixed factors.}
    \label{fig:moe}
\end{figure}

To better understand the FAME architecture, this subsection provides a dedicated explanation of the working mechanism of the gating network within FAME. We choose two tasks: \textbf{Handle Pull} and \textbf{Peg Insert Side}. The gating network is trained using the Gen Dataset ($\mathcal{D}_k$) and the Mix Gen Dataset ($\mathcal{D}_{\text{multi}}$ with $i=2, 3, 4, 5$). 
After training, we feed the observations from the same task or the other task into the model, and then visualize the activation values output by the gating network as heatmaps, as shown in Figure \ref{fig:moe} (we consider 2 tasks so there are $2\times2=4$ visualizations). 
In each subfigure, the horizontal axis represents the number or combination of varying factors, while the vertical axis lists the corresponding factor-specific expert adapters. 

As shown in the first two Figure \ref{subFig:1} and \ref{subFig:2}, when the number of varying factors is small, the gating network tends to focus more on certain specific adapters. As the number of factor variations increases, the activations become more dispersed, reflecting the model's adaptive allocation of experts to handle growing complexity. Notably, as shown in the last two Figure \ref{subFig:3} and \ref{subFig:4}, we also observe that the gate trained on the \textbf{Handle Pull} task can be directly and effectively transferred to the \textbf{Peg Insert Side} task in a zero-shot manner. This cross-task generalization capability suggests that the gating network learns a high-level, task-agnostic representation of visual factors, rather than overfitting to task-specific cues. This further demonstrates the effectiveness of combining adapter network fine-tuning with the MoE architecture.

\subsection{Real World Experiment}

\subsubsection{Experiment Set-up}

\begin{figure}[htbp]
    \centering
    \includegraphics[
        width=0.5\textwidth,
        trim=9cm 6cm 8.5cm 6cm, 
        clip 
    ]{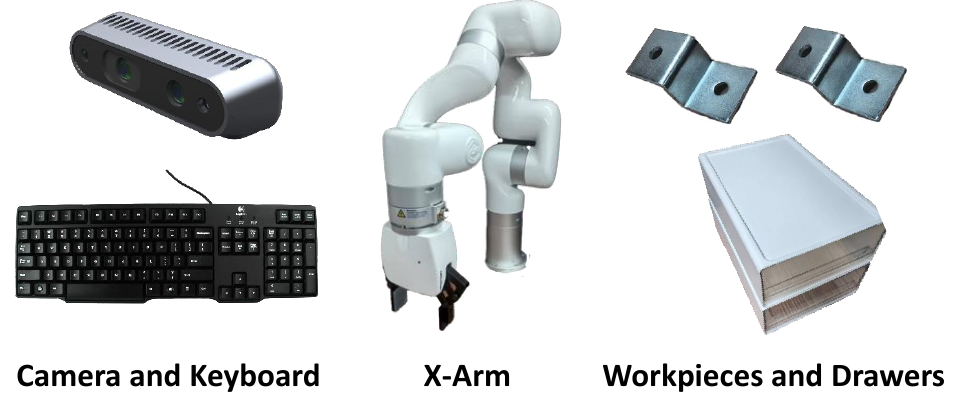}
    \caption{Experimental setup: data collection using keyboard and RealSense camera; X-Arm robotic arm; workpieces and drawer for pick-and-place task.}
    \label{fig:set}
\end{figure}

We conduct all real-world experiments with an X-Arm 6 robotic arm and capture image observations using a RealSense camera. Our focus is on a single pick-and-place task, where the robot needs to pick up workpieces from the tabletop and place them into a drawer. This task requires precise grasping and placement skills in a tabletop manipulation scenario, with detailed setup illustrated in Fig.~\ref{fig:set}.

\begin{figure*}[t]
    \centering
    \includegraphics[width=1.0\linewidth, trim=5mm 10mm 5mm 20mm, clip]{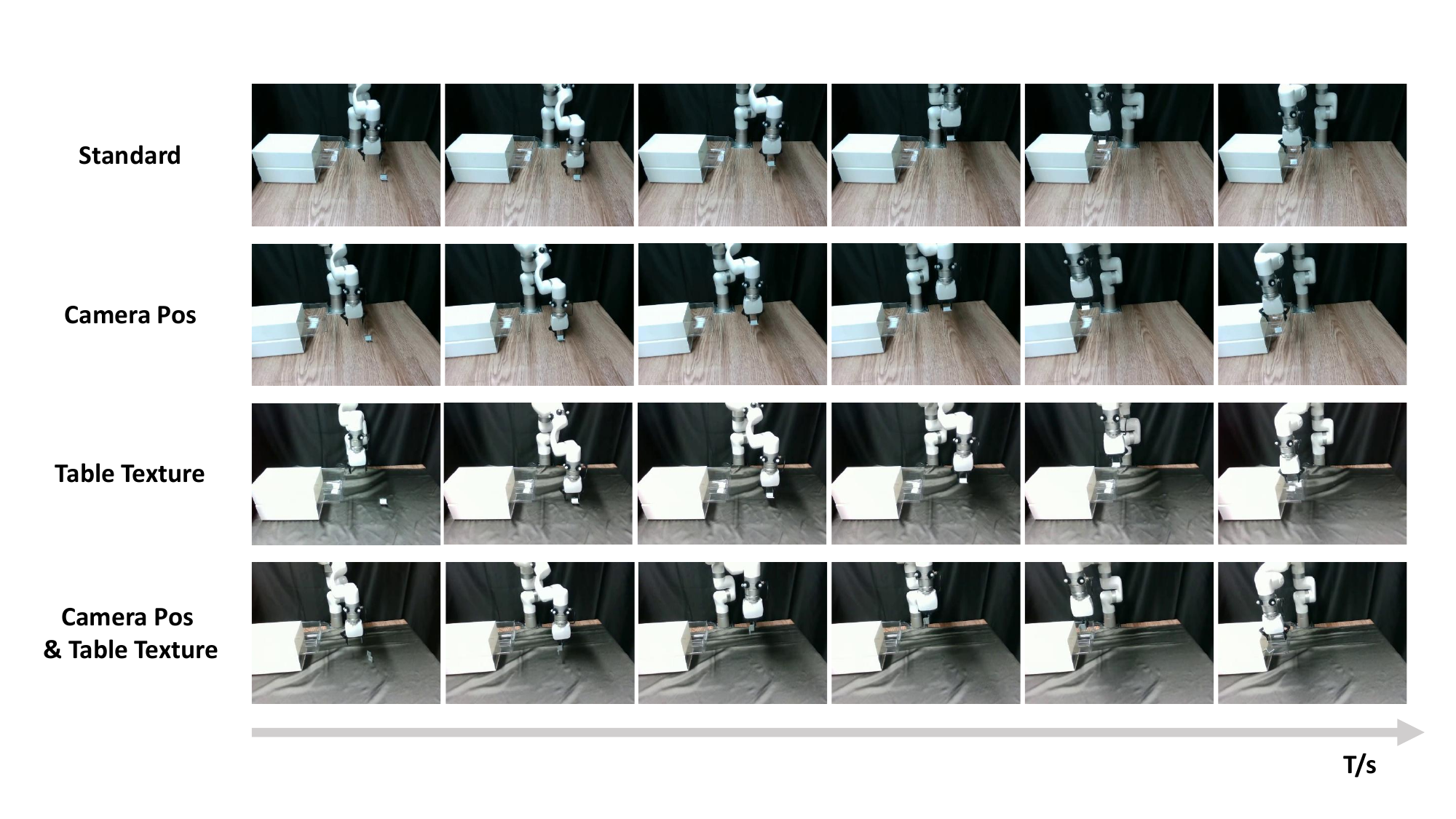}
    \caption{Real-world scenarios for the pick-and-place task: standard environment, camera position changes, table texture changes, and simultaneous changes in camera position and table texture. Keyframes during task execution are captured for each scenario.}
    \label{fig:3}
\end{figure*}

\vspace{0.3em}

\subsubsection{Data Collection}

We collect demonstration data through keyboard teleoperation. For the standard environment of the pick-and-place task, we collect $50$ demonstrations. Additionally, to account for environmental variations, we collect data under two controllable factor changes: (1) camera position changes, and (2) table texture changes. Other factors are evaluated in simulation for safer and more systematic control. For each type of variation, we collect $5$ demonstrations to capture the robot's behavior under different conditions. The real-world scenarios for the pick-and-place task are illustrated in Fig.~\ref{fig:3}.


\vspace{0.3em}

\subsubsection{Training}

We employ knowledge distillation from a pre-trained encoder with Mixture of Experts (MoE) architecture, which was obtained during simulation training on the MetaWorld benchmark, to a smaller ResNet-18 model. We hypothesize that the robust visual representations learned in simulation can transfer to real-world scenarios (sim2real), while the smaller model size facilitates efficient training and deployment in real-world experiments. The distilled model is fine-tuned using the data collected in the previous section. We also analyze the inference latency during real-world deployment. The two-adapter FAME model achieves efficient per-step inference on our hardware and satisfies the real-time control requirements of the X-Arm pick-and-place experiments.

\begin{table}[h]
\centering
\caption{\textbf{Real-world experimental results.} We report ID and OOD success rates under camera-position, table-texture, and combined variations.}
\label{tab:results}
\begin{tabular}{lccccccc}
\toprule
 & \multicolumn{2}{c}{\textbf{Camera Pos}} & \multicolumn{2}{c}{\textbf{Table Texture}} & \multicolumn{2}{c}{\textbf{Camera Pos \&}} & \multirow{2}{*}{\textbf{Average}} \\
 & & & \multicolumn{2}{c}{} & \multicolumn{2}{c}{\textbf{Table Texture}} & \\
\cmidrule(lr){2-3} \cmidrule(lr){4-5} \cmidrule(lr){6-7}
 & ID & OOD & ID & OOD & ID & OOD & \\
\midrule
\textbf{ResNet} & $47$ & $20$ & $40$ & $0$ & $33$ & $20$ & $33$ \\
\textbf{DINO} & $53$ & $0$ & $47$ & $0$ & $40$ & $0$ & $35$ \\
\textbf{FAME} & $\mathbf{87}$ & $\mathbf{60}$ & $\mathbf{80}$ & $\mathbf{40}$ & $\mathbf{67}$ & $\mathbf{40}$ & $\mathbf{70}$ \\
\bottomrule
\end{tabular}
\end{table}

\subsubsection{Evaluation}

For evaluation, we define both In-Distribution (ID) and Out-of-Distribution (OOD) scenarios within three generalization contexts. ID scenarios represent conditions included in the training data distribution, while OOD scenarios represent unseen variations. Specifically, we create OOD conditions by randomly adjusting the initial positions of workpieces and making slight modifications to the relative positions between workpieces and the drawer, ensuring these variations fall outside the training distribution. For each generalization context, the ID scenarios are evaluated 15 times, and the OOD scenarios are evaluated 5 times.

\subsubsection{Main Results}

The real-world experimental results are summarized in Table~\ref{tab:results}. Our FAME framework achieves a superior overall success rate of $70\%$, significantly outperforming the ResNet-DP ($33\%$) and DINO-DP ($35\%$) baselines.

\begin{figure}[htbp]
    \centering
    \includegraphics[
        width=0.5\textwidth,
        trim=8cm 6.5cm 8cm 6cm, 
        clip 
    ]{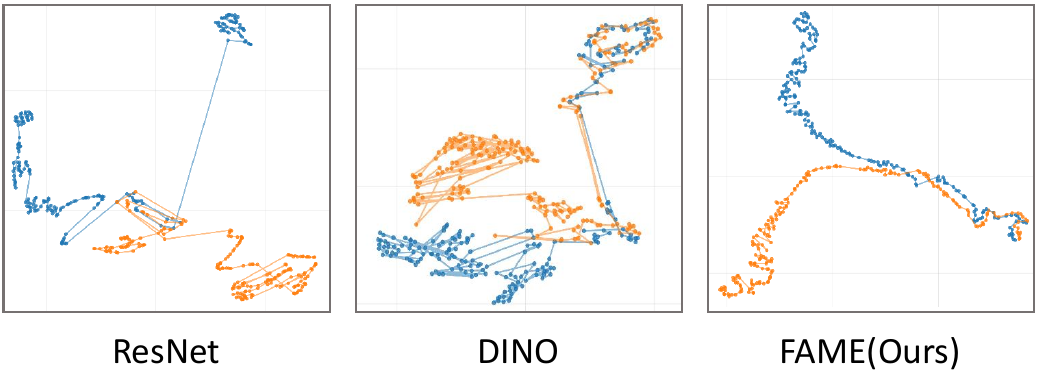}
    \caption{t-SNE visualization of learned feature spaces on the real-world task.}
    \label{fig:visual}
\end{figure}

Notably, FAME demonstrates robust generalization in all tested scenarios. Under camera position variations, it attains success rates of $87\%$ (ID) and $60\%$ (OOD). For table texture changes, it maintains $80\%$ (ID) and $40\%$ (OOD) success. Crucially, in the most challenging combined scenario of simultaneous camera and texture changes, FAME sustains $67\%$ (ID) and $40\%$ (OOD) performance, while baseline methods falter, particularly in OOD conditions.

\subsubsection{Feature Visualization}

We employ t-SNE~\cite{maaten2008visualizing} to visualize the learned feature representations in 2D space. Specifically, we sample frames from two demonstration trajectories collected under different environmental conditions (one trajectory per condition). Each sampled image frame is passed through the respective encoder to obtain its high-dimensional feature representation, which is then projected to 2D via t-SNE for visualization. As shown in Figure~\ref{fig:visual}, each point corresponds to a single frame; points from the same trajectory are colored identically. FAME's features form more structured and well-separated clusters compared to baselines, indicating better organization and disentanglement of environmental factors. This structured feature space directly correlates with FAME's superior generalization capability, as it enables more robust representations under diverse conditions.

\section{CONCLUSION}
\label{sec:conclusion}

\subsubsection{Conclusion}
We proposed FAME, a novel framework that integrates Mixture-of-Experts with frozen pre-trained encoders to enhance combinatorial generalization in robotic manipulation. Our approach features a three-stage training process: (1) policy warm-up with a frozen encoder, (2) factor-specific adapter training on specialized datasets, and (3) joint fine-tuning with a central router that dynamically combines adapters via MoE. This design enables efficient adaptation to multiple environmental variations while preserving valuable pre-trained representations. Extensive experiments on both simulated Meta-World benchmarks and real-world pick-and-place tasks demonstrate FAME's clear superiority over existing methods, achieving significant performance improvements of 34\% in simulation and 35\% in real-world deployment. These results validate FAME's strong generalization capability across diverse and challenging environmental conditions.

\vspace{1mm}

\subsubsection{Limitation and future works}
While FAME shows promising results, our real-world experiments remain limited in scale. Future work could validate more real-world factor combinations, explore efficient adapter architectures, and investigate factor correlations in greater depth.

\bibliographystyle{IEEEtran} 
\bibliography{IEEEexample}

\end{document}